\documentclass[10pt,twocolumn,letterpaper]{article}

\usepackage[pagenumbers]{cvpr}
\usepackage{fancyhdr}
\usepackage{setspace}

\input{preamble}

\usepackage{pgfplots}
\makeatletter
\@ifpackagelater{pgfplots}{2021/05/14}
  {\pgfplotsset{compat=1.18}}
  {\pgfplotsset{compat=1.17}}
\makeatother

\definecolor{cvprblue}{rgb}{0.21,0.49,0.74}
\usepackage[pagebackref,breaklinks,colorlinks,allcolors=cvprblue]{hyperref}

\AtEndPreamble{
    \usepackage[capitalize]{cleveref}
    \crefname{section}{Sec.}{Secs.}
    \Crefname{section}{Section}{Sections}
    \Crefname{table}{Table}{Tables}
    \crefname{table}{Tab.}{Tabs.}
}

\makeatletter
\let\arxiv@cvpr@maketitle\@maketitle
\newcommand{\arxivsettitle}[1]{\gdef\@title{#1}\gdef\thetitle{#1}}
\newcommand{\arxivsetauthor}[1]{\gdef\@author{#1}}
\newcommand{\arxivmaketitle}{%
  \if@twocolumn
    \ifnum \col@number=\@ne
      \arxiv@cvpr@maketitle
    \else
      \twocolumn[\arxiv@cvpr@maketitle]%
    \fi
  \else
    \newpage
    \global\@topnum\z@
    \arxiv@cvpr@maketitle
  \fi
}
\newcommand{\arxivaliascite}[2]{%
  \@ifundefined{b@#2}{}{%
    \expandafter\let\expandafter\arxiv@sourcecite\csname b@#2\endcsname
    \global\expandafter\let\csname b@#1\endcsname\arxiv@sourcecite
  }%
}
\newcommand{\arxivciteanchor}[1]{%
  \Hy@raisedlink{\hyper@anchorstart{cite.#1}\hyper@anchorend}%
}
\let\arxiv@NAT@sort@cites\NAT@sort@cites
\newcommand{\arxivstopcitationwrites}{%
  \gdef\NAT@sort@cites##1{%
    \let\NAT@cite@list\@empty
    \@for\@citeb:=##1\do{\expandafter\NAT@star@cite\@citeb\@@}%
    \@ifnum{\NAT@sort>\z@}{%
      \expandafter\NAT@sort@cites@\expandafter{\NAT@cite@list}%
    }{}%
  }%
}
\AtBeginDocument{%
  \arxivaliascite{main-He:2022:MAA}{He:2022:MAA}%
  \arxivaliascite{main-Simeoni:2025:DINOv3}{Simeoni:2025:DINOv3}%
  \arxivaliascite{main-Araslanov:2020:SSS}{Araslanov:2020:SSS}%
  \arxivaliascite{main-Oquab:2023:DINOv2}{Oquab:2023:DINOv2}%
  \arxivaliascite{main-Radford:2021:LTV}{Radford:2021:LTV}%
}
\makeatother

\newcommand{\arxivauthors}{Nikita Araslanov$^{1\,2\,3}$ \qquad Martin Sundermeyer$^1$ \qquad Hidenobu Matsuki$^1$ \\[1mm] David Joseph Tan$^1$ \quad Federico Tombari$^{1\,2\,3}$ \\[3mm]
$^1$\,Google \qquad $^2$\,TU Munich \qquad $^3$\,Munich Center for Machine Learning\\
}

\begin{document}

\pagenumbering{arabic}

\arxivsettitle{Featurising Pixels from Dynamic 3D Scenes with Linear In-Context Learners}
\arxivsetauthor{\arxivauthors}
\arxivmaketitle
\thispagestyle{fancy}

\begin{abstract}
One of the most exciting applications of vision models involve pixel-level reasoning.
Despite the abundance of vision foundation models, we still lack representations that effectively embed spatio-temporal properties of visual scenes at the pixel level.
Existing frameworks either train on image-based pretext tasks, which do not account for dynamic elements, or on video sequences for action-level reasoning, which does not scale to dense pixel-level prediction.
We present a framework that learns pixel-accurate feature descriptors from videos, \oursName.
The core element of our training framework is linear in-context learning.
\oursName leverages spatio-temporal cue maps\emdash depth and motion\emdash estimated with off-the-shelf networks.
Despite the noisy nature of those cues, \oursName trains effectively on uncurated video datasets, embedding semantic and geometric properties in a temporally consistent manner.
We demonstrate compelling empirical benefits of the learned representation across a diverse suite of vision tasks: video object segmentation, surface normal estimation and semantic segmentation.\projectwebsite{https://lila-pixels.github.io}
\end{abstract}

\begin{figure}[t]
    \centering
    \def\svgwidth{1.0\linewidth}
    \input{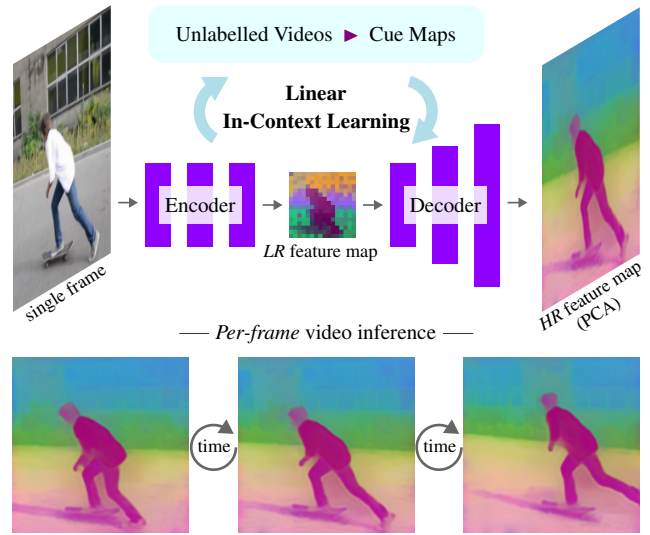}
    \caption{\textbf{Overview.} We train an encoder-decoder model from unlabelled videos to produce high-resolution (HR), temporally consistent feature maps. The core novelty of our training approach is \emph{linear in-context learning}, or \emph{\oursName}. Trained on noisy cue maps, such as those provided by off-the-shelf optical flow and monocular depth networks, \oursName enhances the low-resolution (LR) encoder features with pixel-level geometry and temporally stable semantics.}
    \label{fig:teaser}
\end{figure}

\section{Introduction}

Most computer vision tasks require pixel-level feature representations that embed both semantic and geometric properties of a visual scene.
Although state-of-the-art foundation models, such as the DINO family \cite{Oquab:2023:DINOv2,Caron:2021:EPS,Simeoni:2025:DINOv3}, already encode a remarkable degree of these properties, these encoder-only models yield feature maps at a low \emph{patch}-level resolution.
To achieve \emph{pixel}-fine feature grids directly, one could upsample the input by a factor of the patch size.
However, this approach is neither computationally efficient, nor task-effective in practice, since it introduces a discrepancy in input resolution between training and inference.
In this work, we introduce a novel training strategy for encoder-decoder architectures, which natively computes a feature vector for each input pixel, as shown in \cref{fig:teaser} (top).

The core novelty of our work is \emph{linear in-context learning}.
The idea is to learn a representation that remains \emph{consistent} across a video sequence under a linear projection, which maps the representation to a set of \emph{cue maps}.
Specifically, an optimal linear projection for mapping a feature grid of one frame to a cue (\eg a depth map), must also produce the corresponding cue (\ie a depth map) for the other frames in the sequence. 
In this work, we consider two cue modalities\emdash depth and optical flow\emdash which we complement with self-distillation of upsampled encoder representations.
\textbf{L}inear \textbf{i}n-context \textbf{l}e\textbf{a}rning (\emph{\oursName}) leads to encoder-decoder foundation models that effectively embed spatio-temporal properties of visual scenes, encompassing geometric and semantic cues with strong temporal consistency (see \cref{fig:teaser}, bottom).

In contrast to prior work on feature upsampling \cite[\eg][]{Huang:2025:LoftUp,Fu:2024:FeatUp}, we harness video datasets for training \oursName.
The abundance of videos promises feature representations with properties complementary to image-based pre-training (\eg motion cues \cite{Araslanov:2025:FFT}).
Off-the-shelf networks for estimating depth and optical flow already exhibit impressive generalisation to in-the-wild videos.
Since the training data for these networks typically comprises a mixture of synthetic and real-world data captured with inexpensive hardware (\eg RGB-D cameras),
the overall annotation effort is substantially lower compared to methods using mask supervision \cite{Huang:2025:LoftUp,Kirillov:2023:SAM}.
However, a crucial challenge in learning from the cues predicted by pre-trained models is the inherent noise and imperfections in the training signal.
We show that \oursName learns effectively despite the imperfect cue maps.
Although trained on videos, \oursName requires only a single image as the input at inference.
This sets \oursName apart from the prior work on video-based pre-training that operates on spatio-temporal inputs both at training and inference \cite[\eg][]{tong:2022:videomae,Bardes:2024:RFP}.

Our work makes two core contributions.
\textit{(i)} We develop a novel training technique, linear in-context learning, for encoder-decoder pre-training from videos.
Our model yields dense feature maps per image with powerful geometric, semantic and temporal qualities.
\textit{(ii)} We rigorously evaluate \oursName on a diverse suite of tasks \emph{distinct} from the cue modalities used during training.
Specifically, \oursName yields consistent and significant improvement over established baselines across three benchmarks: video object segmentation, surface normal estimation and semantic segmentation.

\section{Related work}

Our work contributes to representation learning from unlabelled videos.
Therefore, \oursName is related to the literature of self- and weakly supervised learning.

\inparagraph{Self- and weakly supervised learning.}
The most widely used representations stem from pre-training on image data \citep{Oquab:2023:DINOv2,He:2022:MAA}.
\textit{Contrastive learning} \citep{Gutmann:2010:NCE} maximises the similarity of a sample to its positive counterpart \wrt a set of negative samples \citep{Oord:2018:InfoNCE}, and underpins a large family of frameworks \citep{Chen:2020:SimCLR,Chen:2021:MoCoV3,Chen:2021:ESS}. For example, MoCo \citep{Chen:2021:MoCoV3} applies the contrastive loss by deriving the negative samples from a dynamic queue.
Non-contrastive methods, such as DINO \citep{Caron:2021:EPS} and its scaled-up variants \citep{Oquab:2023:DINOv2,Simeoni:2025:DINOv3}, implement clustering via self-distillation, where the mean teacher creates categorical targets for the student.
BYOL \citep{Grill:2020:BYOL} is another example of a non-contrastive approach with an asymmetric architecture, which dispenses with the negative samples without solution collapse.
Masked autoencoder \citep[MAE,][]{He:2022:MAA} implements \textit{masked image modelling (MIM)}, \ie learning to reconstruct the input from its partial observation. 
Instead of the colour space as in MAE, I-JEPA \citep{Assran:2023:IJEPA} realises MIM in the embedding space, by predicting the \emph{representation} of masked patches.
BEiT \citep{Bao:2022:BEiT} and iBOT \citep{Zhou:2022:iBOT} leverage a pre-trained tokeniser to encode the targets for the partially masked input.
Training on paired text-image data, CLIP \citep{Radford:2021:LTV} and SigLIP \citep{Zhai:2023:SigLIP,Tschannen:2025:SigLIPv2} produce semantically aligned representations by learning a joint embedding space shared by the two modalities.
Given the diversity of foundation models, feature aggregators, such as AM-RADIO \citep{Ranzinger:2024:AMRADIO}, aim at creating ``superset'' models by distilling multiple foundation models into one.

A practical limitation of the works above is the low resolution of the feature grid, comprising only \textit{patch} features.
While there are upsampling strategies, such as FeatUp \citep{Fu:2024:FeatUp} and LoftUp \citep{Huang:2025:LoftUp}, we take an approach that yields \emph{pixel-level} representations naturally and \emph{complementary} (\ie with different properties) to those obtained from the encoder.

\inparagraph{Dense (pixel-level) representation learning.}
Previous work has approached dense representations with pixel-level contrastive learning \citep{Pinheiro:2020:ULD} and clustering with multi-crop \citep{Henaff:2022:ODR} and cross-image consistency \citep{Lebailly:2024:CrIBo}.
Other dense pre-text tasks include creating targets for relative position prediction (either for patches \cite{Noroozi:2016:ULV} or entire crops \citep{Caron:2024:LOCA}), and colorisation \citep{Zhang:2016:CIC,Larsson:2017:CPT}.
CroCo \citep{Weinzaepfel:2022:CroCo} learns to complete the second view given the context of an adjacent view.

\begin{figure*}[t]
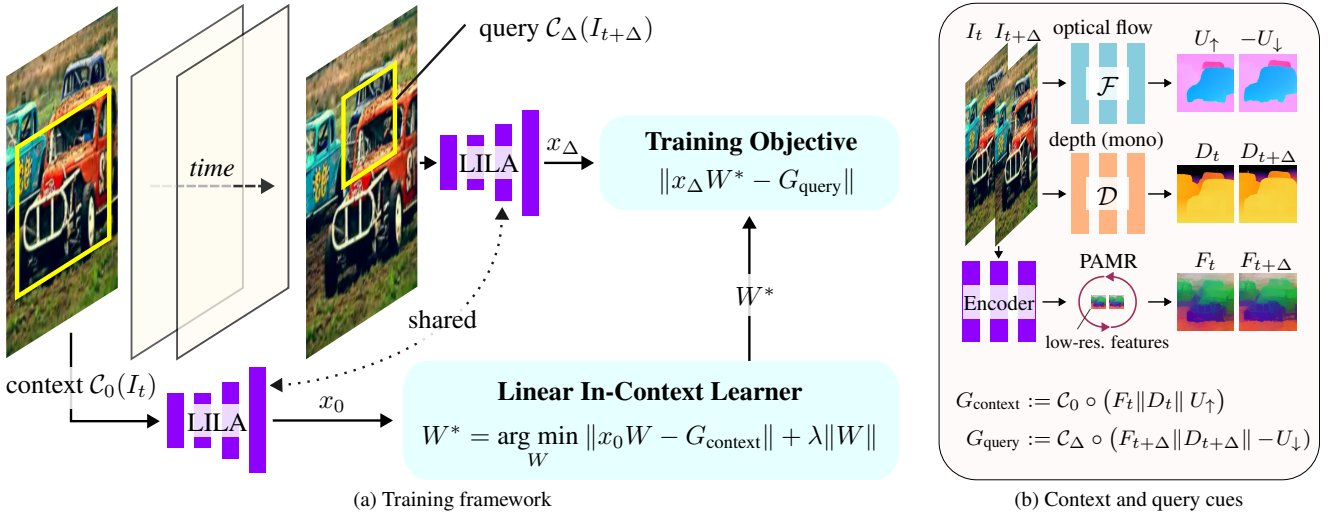

    \centering
    
    \begin{subfigure}[b]{0.68\textwidth}
        \centering
        \def\svgwidth{1.0\linewidth}
        \input{figures/arch/arch_v3_a_overlay.tex}
        \subcaption{Training framework}
        \label{fig:arch_framework}
    \end{subfigure}
    \hfill
    \begin{subfigure}[b]{0.29\textwidth}
        \centering
        \def\svgwidth{1.0\linewidth}
        \input{figures/arch/arch_v3_b_overlay.tex}
        \subcaption{Context and query cues}
        \label{fig:arch_cues}
    \end{subfigure}
    
    \caption{\textbf{Training overview.} \subref{fig:arch_framework} We train an encoder-decoder network (\oursName) to produce pixel-level feature maps (here, $x_\cxt$ and $x_\qry$). As the pretext task, we propose \emph{linear in-context learning}. Solving for a linear projection $W^\ast$ \wrt \emph{context} frame $I_t$ and the corresponding cue maps $G_\text{context}$ (see \subref{fig:arch_cues}), we train the network to minimise the reconstruction loss \wrt the features produced for a \emph{query} frame $I_{t + \Delta}$ and the cues $G_\text{query}$ under the \emph{same} linear projection $W^\ast$ derived from the context.}
    \label{fig:arch}
\end{figure*}

\inparagraph{Dense representation learning from videos.}
Videos are abundant, but learning dense representations from videos has proved challenging.
Jabri~\etal~\cite{Jabri:2020:CRW} combine contrastive learning with cycle consistency by propagating a patch-level label set forward and backward.
Araslanov~\etal~\cite{Araslanov:2021:DUL} create the learning signal with a batch-level set of clusters spread uniformly across multiple videos.
VITO \citep{Parthasarathy:2023:SSV} uses attention to spatially aggregate a representation most likely predictive of the future state.
Leveraging pre-trained optical flow networks, FlowE \cite{Xiong:2021:FlowE} extends the BYOL framework to pixel-level training.
Specifically, FlowE encourages representation similarity of temporally corresponding pixels. 
PooDLe \cite{Wang:2025:PooDLe} augments the dense objective of FlowE with temporally coherent crop pairs.
DORA~\citep{Venkataramanan:2024:DORA} uses a student-teacher framework by distilling object-patch correspondences.
Leveraging slot attention and sparse depth supervision, SAVi++ \citep{Elsayed:2022:SAVi} encodes a video into a set of temporally invariant slots.

In contrast to these works, we explore geometrically grounded signals for supervision\emdash depth and motion\emdash with a simple novel technique, \emph{linear in-context learning}.

\section{Linear In-Context Learning}

\inparagraph{Overview.}
\cref{fig:arch} provides an illustrative overview of linear in-context learning (\oursName).
The training operates on a video dataset.
Each training iteration takes a batch of adjacent frame pairs $(I_t, I_{t+\Delta})$ at resolution $N := H \times W$, sampled with a varying temporal window.
We designate frame $I_t$ as the \emph{context} frame, and frame $I_{t+\Delta}$ as the \emph{query}.
The key idea is that the optimal linear mapping $W^\ast$ from the context representation to the corresponding cues must also apply for the representation of the query frame.
In this work, we apply LILA to the cue maps combining two modalities\emdash depth and optical flow\emdash estimated from pre-trained networks.

We bootstrap our model from a pre-trained vision transformer, such as DINOv2 \cite{Bhat:2023:ZDZ}, and equip the backbone with a DPT decoder \citep{Ranftl:2021:VTD}.
The DPT decoder has skip connections to four intermediate feature blocks in the encoder, which allows it to effectively upsample the patch-level tokens to a pixel-level feature map.
Our framework trains only the decoder, while keeping the encoder parameters frozen.

While we estimate the depth and flow maps using frames $(I_t, I_{t + \Delta})$, we provide only their crops as the input to our encoder-decoder network.  
We randomly generate aspect-preserving crops $c_\cxt := \mathcal{C}_\cxt(I_t)$ and $c_\qry := \mathcal{C}_\qry (I_{t + \Delta})$, and feed them to \oursName in parallel.
We aim to train our network to produce useful pixel-dense representations by imposing a training loss on the corresponding outputs $x_\cxt, x_\qry \in \mathbb{R}^{N \times d}$.

\begin{figure*}[t]
    \centering
    
    \begin{subfigure}[t]{0.135\textwidth}
        \centering
        \FixedHeader{\emph{Context} Input $I_t$} \\
        \includegraphics[width=\linewidth]{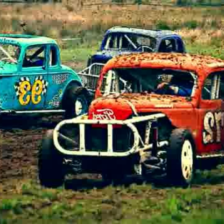}
    \end{subfigure}%
    \hfill
    \begin{minipage}[t]{0.28\textwidth}
        \centering
        \FixedHeader{\emph{Query} Flow $U_{t + \Delta}$} \\
        
        \begin{overpic}[width=0.48\linewidth]{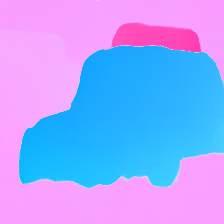}
            \put(3,5){\scriptsize\color{white}\textbf{Cue}}
        \end{overpic}%
        \hspace{0em}
        \begin{overpic}[width=0.48\linewidth]{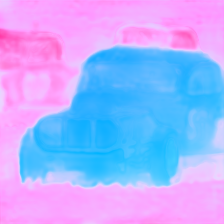}
            \put(3,5){\scriptsize\color{white}\textbf{Prediction}}
        \end{overpic}
    \end{minipage}%
    \hfill
    \begin{minipage}[t]{0.28\textwidth}
        \centering
        \FixedHeader{\emph{Query} Depth $D_{t + \Delta}$} \\
        
        \begin{overpic}[width=0.48\linewidth]{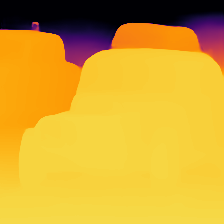}
            \put(3,5){\scriptsize\color{black}\textbf{Cue}}
        \end{overpic}%
        \hspace{0em}
        \begin{overpic}[width=0.48\linewidth]{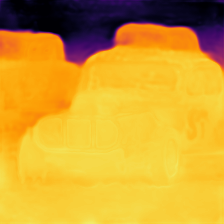}
            \put(3,5){\scriptsize\color{black}\textbf{Prediction}}
        \end{overpic}%
    \end{minipage}%
    \hfill
    \begin{minipage}[t]{0.28\textwidth}
        \centering
        \FixedHeader{\emph{Query} Features $F_{t + \Delta}$} \\
        
        \begin{overpic}[width=0.48\linewidth]{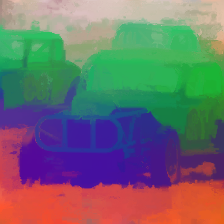}
            \put(3,5){\scriptsize\color{white}\textbf{Cue}}
        \end{overpic}%
        \hspace{0em}
        \begin{overpic}[width=0.48\linewidth]{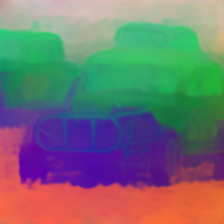}
            \put(3,5){\scriptsize\color{white}\textbf{Prediction}}
        \end{overpic}%
    \end{minipage}

    \caption{\textbf{Approximating cue maps.} \oursName learns to approximate the query cues (flow, depth and encoder features) with high fidelity in a temporally consistent fashion. Providing a glimpse into the \oursName's training process, the predictions here are derived from an optimal linear mapping \wrt the \emph{context} frame that is distinct from the query frame, thereby enforcing spatio-temporal coherence.}
    \label{fig:lila_approx}
    \vspace{-0.5em}
\end{figure*}

\inparagraph{Cue maps: motion, depth and self-distillation.}
To supervise the network, we use pixel-level cue maps, illustrated in \cref{fig:arch_cues}.
Specifically, we use optical flow computed between $(I_t, I_{t+\Delta})$ in the forward $U_\uparrow$ and backward directions $U_\downarrow$, and depth estimates $D_t$ and $D_{t+\Delta}$ for frames $I_t$ and $I_{t+\Delta}$.
Note that $U_\uparrow$ is defined in the reference frame of $I_t$, warping it to frame $I_{t+\Delta}$ (conversely for $U_\downarrow$).
Furthermore, we complement the flow and depth with low-resolution encoder features.
We refine the features with the image statistics using \emph{pixel-adaptive map refinement} (PAMR) \cite{Araslanov:2020:SSS}.
PAMR implements a variant of mean-field inference for a Conditional Random Field (CRF) using local affinity-based kernels.
As PAMR is computationally efficient, it does not cause a significant training overhead.\footnote{In practice, we also use the encoder features already available from the forward pass of the inputs $c_\cxt$ and $c_\qry$, removing the need for extra inference.}
We adapt PAMR to refine the encoder's feature grids in a coarse-to-fine fashion (see the \supp for details).
Let us denote the feature maps refined by PAMR as $F_t$ and $F_{t+\Delta}$.

To construct the cue maps, we concatenate the optical flow, the depth maps and the refined encoder features.
Formally, we define the target cue maps as
\begin{equation}\label{eq:g_naive}
\begin{aligned}
    G_\text{context} &:= \mathcal{C}_\cxt \circ (F_t \concat D_t \concat U_\uparrow), \\
    G_\text{query} &:= \mathcal{C}_\qry \circ (F_{t + \Delta} \concat D_{t + \Delta} \concat -U_\downarrow),
\end{aligned}
\end{equation}
where operator $\concat$ denotes concatenation along the feature dimensionality.
We apply the corresponding crops $\mathcal{C}_\cxt$ and $\mathcal{C}_\qry$ to align the cue maps with the feature grids of the \oursName network.
Note also that $U_\downarrow$ is the \emph{backward} flow, thus we invert its sign for consistency with $U_\uparrow$.

\inparagraph{Training \oursName.}
Using the cue maps $G_\text{context}, G_\text{query} \in \mathbb{R}^{N \times m}$, we develop an in-context training objective.
Concretely, we aim to obtain the feature maps $x_\cxt, x_\qry$ such that an optimal linear projection of $x_\cxt$ approximating $G_\text{context}$ also works for $x_\qry$ to approximate $G_\text{query}$, \ie network predictions remain \emph{consistent} under the linear transformation.
More formally, for depth maps and encoder features, the consistency implies that if a linear operator $W$ optimally approximates these cues for frame $I_t$, it must approximate them for frame $I_{t+\Delta}$. 
For optical flow, the consistency implies that if an operator $W$ approximates the forward flow, $-W$ must approximate the backward flow.
We formulate this idea into a learning framework.

Given $x_\cxt$, produced by our encoder-decoder model for crop $\mathcal{C}_\cxt(I_t)$, we find the optimal projection $W^\ast$ \wrt $G_\text{context}$ via linear ridge regression:
\begin{equation}\label{eq:lstsq_obj}
    W^\ast = \underset{W}{\text{arg min}}\:\| x_\cxt W - G_\text{context}\| + \lambda \| W \|.
\end{equation}
Note that \cref{eq:lstsq_obj} has a closed-form solution and is efficient to compute.
Specifically, the problem has the size $d$, which is the feature dimensionality, not $N$, the number of pixels \citep{Araslanov:2025:FFT}.
Next, using the optimal $W^\ast$, we compute the loss by minimising the distance between the feature map $x_\qry$ and the corresponding encoding $G_\text{query}$ for a cropped adjacent frame $\mathcal{C}_\qry(I_{t + \Delta})$:
\begin{equation}\label{eq:cycle-loss}
    \mathcal{L}_\text{L1} = \| x_\qry W^\ast - G_\text{query}\|_1.
\end{equation}

Discontinuities in the cue maps are highly correlated with the semantic structure of the scene.
Therefore, we additionally encourage edge consistency with a gradient matching loss \cite{Araslanov:2025:FFT}, defined for the horizontal axis as
\begin{equation}\label{eq:grad-loss}
    \mathcal{L}_{\nabla\texttt{x}} = \omega_\texttt{x} \: \|  \nabla_\texttt{x}(x_\qry W^\ast) - \nabla_\texttt{x} G_\text{query} \|_1,
\end{equation}
where $\omega_\texttt{x} = 1 - \exp{(-\nabla_\texttt{x} G_\text{query} / \sigma)}$ and the operator $\nabla_\texttt{x}$ takes the absolute value of the gradient.
Equivalently, we define $\mathcal{L}_{\nabla,\texttt{y}}$ for the vertical axis.
Combining the losses, 
\begin{equation}
    \mathcal{L}_\nabla = \mathcal{L}_{\nabla\texttt{x}} + \mathcal{L}_{\nabla\texttt{y}},
\end{equation}
we compute the total loss for training \oursName:  
\begin{equation}
    \mathcal{L}_\text{\oursName} = \mathcal{L}_\text{L1} + \gamma  \mathcal{L}_\nabla.
\end{equation}

\inparagraph{Discussion.} 
Linear in-context learning imposes a non-trivial training task on the encoder-decoder network.
The optimal linear map $W^\ast$, derived via ridge regression, must capture the visual elements of the context frame that remain \emph{invariant} across the \emph{entire} temporal window.
The modalities embedded by the cue maps further encourage the network to learn a spatially diverse structure, specifically recognising  dynamic objects (due to optical flow) and encoding scene geometry (due to depth maps).
By embedding the refined encoder maps, \oursName simultaneously performs a variant of \emph{self-distillation}, which preserves the encoder's original semantic properties.
\cref{fig:lila_approx} visualises the target cues and their approximations during training of \oursName.
Despite the inherent noise and inaccuracies present in all cue modalities, \oursName learns an effective representation, as we demonstrate experimentally in the next section.

\inparagraph{Inference.}
At test time, \oursName operates in a standard fashion, discarding the depth and optical flow networks.
It accepts a single image as input and produces a feature map of the same spatial resolution.
\section{Experiments}

\inparagraph{Implementation details}
We train the \oursName models with AdamW \citep{Loshchilov:2019:AdamW} using the learning rate $10^{-4}$ and weight decay $10^{-5}$.
The decoder architecture is based on DPT \citep{Ranftl:2021:VTD}.
We define the output feature dimensionality from the decoder as $128$, $192$ and $256$ for the \oursName variants with backbones ViT-S14, ViT-B14, ViT-L14, respectively.
We set $\gamma = 1$ throughout all experiments; \oursName is not sensitive to moderate deviations of this hyperparameter.
For the experiments below, we initialise the backbones from DINOv2~\cite{Oquab:2023:DINOv2}.
However, we show that \oursName generalises effectively to backbones with other pre-training strategies (see the \supp).

\inparagraph{Training datasets.}
We experiment with two datasets for training \oursName: YouTube-VOS (YT-VOS) \cite{Xu:2018:YTVOS} and Kinetics-700 \cite{Kay:2017:Kinetics}.
The YouTube-VOS (2019) dataset comprises $3,978$ (\textit{train} and \textit{val}) high-resolution video sequences of everyday scenes.
Kinetics-700 contains up to $650$K video clips of human actions.
We do not use any annotation accompanying these datasets.
We train the smaller \oursName variants on YouTube-VOS (ViT-S14, ViT-B14), while for the larger backbone (VIT-L14) we experiment with both training sets.

\inparagraph{Evaluation datasets.}
We test \oursName on three distinct tasks: video object segmentation (VOS), surface normal estimation and semantic segmentation.
DAVIS-2017~\cite{PontTuset:2017:DAV} is the standard semi-supervised VOS benchmark with $30$ video sequences in the validation set.
For semantic segmentation, we use COCO-Stuff~\cite{Caesar:2018:COCO} with 27 coarse categories.
NYUv2~\cite{Silberman:2012:ECCV12} provides the testbed for surface normal estimation, focusing primarily on indoor environments.

\begin{table*}[t]
\newcolumntype{Y}{S[table-format=2.1]@{\hspace{1.5em}}} 
\newcolumntype{Z}{S[table-format=2.1]} 
\medskip
\footnotesize
\centering
\captionof{table}{\textbf{Video object segmentation (DAVIS-2017, val).} We evaluate \oursName with linear probing and local \textit{k}-NN, and report the mean Jaccard index $\mathcal{J}_m$, the boundary F-Score $\mathcal{F}_m$ and their mean $\mathcal{JF}$. Across both probing methodologies and all metrics, \oursName outperforms the baselines and previous work by a significant margin.}
\label{exp:vos}
\begin{tabularx}{\linewidth}{Xcp{8em}@{}YYZp{5em}@{}YYZ}
\toprule
\multirow{2}{*}{Method} & \multirow{2}{*}{Train Data} & & \multicolumn{3}{c}{Linear Probing} & & \multicolumn{3}{c}{Local \textit{k}-NN} \\
\cmidrule(lr){4-6} \cmidrule(lr){8-10}
& & & $\mathcal{JF}$ & $\mathcal{J}_m$ & $\mathcal{F}_m$ & & $\mathcal{JF}$ & $\mathcal{J}_m$ & $\mathcal{F}_m$ \\
\midrule
V-JEPA~\cite{Bardes:2024:RFP} & VideoMix2M~\cite{Bardes:2024:RFP}    & & 49.0 & 46.1 & 51.9  & & 56.7 & 55.6 & 57.8 \\
VideoMAE~\citep{tong:2022:videomae} & Kinetics    & & 43.3 & 40.9 & 45.8 & & 55.1 & 54.6 & 55.6 \\
CRW-ResNet18~\cite{Jabri:2020:CRW}  & Kinetics    & & {\textit{n/a}} & {\textit{n/a}} & {\textit{n/a}} & & 67.6 & 64.8 & 70.2 \\
\midrule
DINO2-S14 \citep{Oquab:2023:DINOv2} &    LVD$^\ast$ & & 57.5 & 54.2 & 60.7 & & 65.1 & 63.7 & 66.6 \\
\leftPad FeatUp ~\citep{Fu:2024:FeatUp}   & \quad \:\leftPad COCO-S    & & 60.5 & 57.4 & 63.6 & & 65.5 & 65.0 & 66.1 \\
\leftPad LoftUp~\citep{Huang:2025:LoftUp}        & \quad \!\quad \leftPad   SA1B \cite{Kirillov:2023:SAM}     & & 63.0 & 59.6 & 66.4 &  & 66.0  & 64.7  & 67.4 \\ 
\leftPad FlowFeat \citep{Araslanov:2025:FFT}               & \quad \leftPad  YT-VOS     & & 65.8 & 62.0 & 69.7 & & 67.6 & 65.6 & 69.6 \\
\rowcolor{azure}
\leftPad \oursName (ours)               & \quad \leftPad  YT-VOS    & & \bfseries 68.6 & \bfseries 66.0 & \bfseries 71.2 & & \bfseries 73.9 & \bfseries 71.7 & \bfseries 76.0 \\ 
\midrule
DINO2-B14 \citep{Oquab:2023:DINOv2} &              LVD$^\ast$        & & 61.6 & 58.5 & 64.7 & & 66.4 & 64.4 & 68.5 \\
\leftPad FlowFeat \citep{Araslanov:2025:FFT}                & \quad \leftPad  YT-VOS     & & 65.7 & 62.2 & 69.2 & & 69.0 &  66.9 & 71.2 \\ 
\rowcolor{azure}
\leftPad \oursName (ours)                & \quad \leftPad  YT-VOS     & & \bfseries 70.4 & \bfseries 67.7 & \bfseries 73.0 & & \bfseries 74.2 & \bfseries 71.8 & \bfseries 76.6 \\ 
\midrule
DINO2-L14 \cite{Oquab:2023:DINOv2} &  LVD$^\ast$        & & 61.0 & 57.9 & 64.0 & & 65.5 & 64.8 & 66.3 \\
\leftPad FlowFeat \citep{Araslanov:2025:FFT}               & \quad \leftPad  YT-VOS     & & 66.9 & 63.4 & 70.4 & & {\textit{n/a}} & {\textit{n/a}} & {\textit{n/a}} \\
\rowcolor{azure}
\leftPad \oursName (ours)               & \quad  \leftPad  YT-VOS     & & 69.0 & 66.0 & 72.0 & & 74.7 & 72.2 & 77.1 \\
\rowcolor{azure}
\leftPad \oursName (ours)               & \quad \!\leftPad  Kinetics  & & \bfseries 69.3 & \bfseries 66.1 & \bfseries 72.5 & & \bfseries 74.9 & \bfseries 72.5 & \bfseries 77.2 \\
\bottomrule
\end{tabularx}
\vspace{-0.8em}
\end{table*}

\begin{figure*}[t]
    \centering

    \begin{subfigure}[t]{\textwidth}
        \centering
        \begin{minipage}{0.195\linewidth}\centering\footnotesize Input\end{minipage}\hfill
        \begin{minipage}{0.195\linewidth}\centering\footnotesize DINOv2 (PCA)\end{minipage}\hfill
        \begin{minipage}{0.195\linewidth}\centering\footnotesize DINOv2 (Prediction)\end{minipage}\hfill
        \begin{minipage}{0.195\linewidth}\centering\footnotesize \oursName\ (PCA)\end{minipage}\hfill
        \begin{minipage}{0.195\linewidth}\centering\footnotesize \oursName\ (Prediction)\end{minipage}

        \begin{overpic}[width=0.195\linewidth]{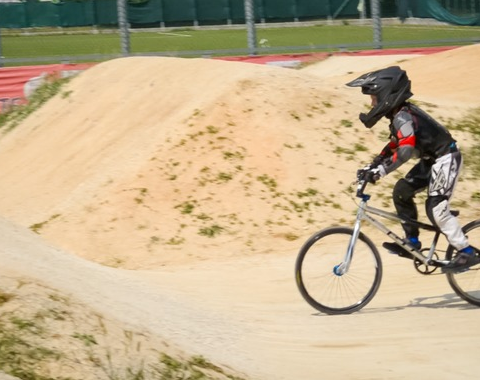}
            \put(75,70){\scriptsize\color{white}\textbf{$t = 1$}}
        \end{overpic}\hfill
        \includegraphics[width=0.195\linewidth]{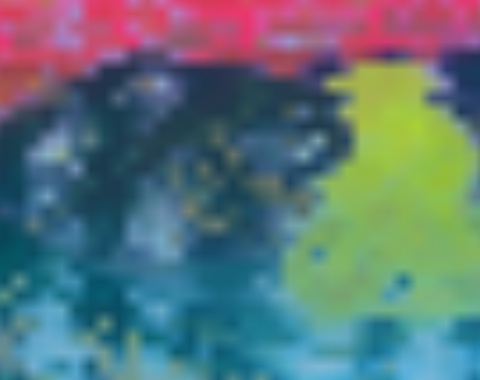}\hfill
        \includegraphics[width=0.195\linewidth]{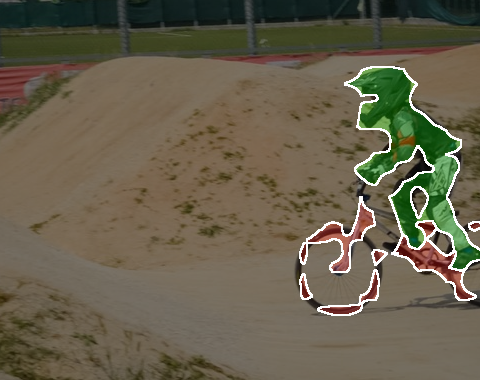}\hfill
        \includegraphics[width=0.195\linewidth]{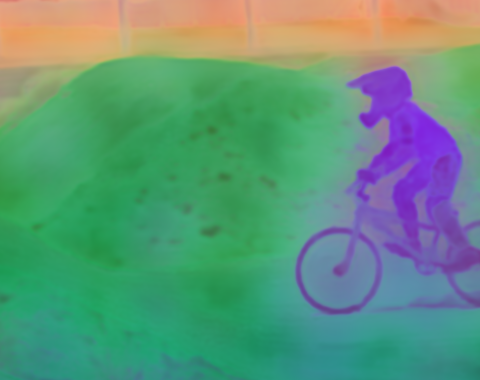}\hfill
        \includegraphics[width=0.195\linewidth]{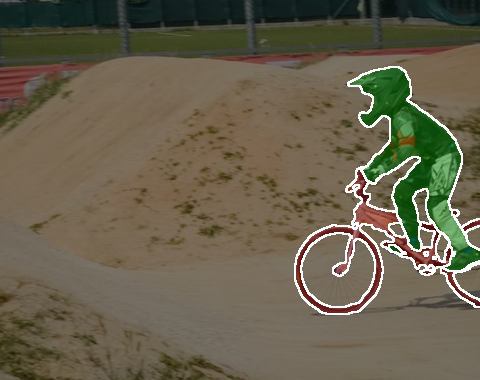}
    \end{subfigure}%

    \vspace{0.2em}

    \begin{subfigure}[t]{\textwidth}
        \centering

        \begin{overpic}[width=0.195\linewidth]{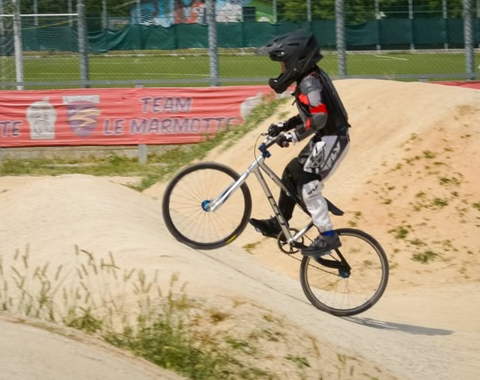}
            \put(75,70){\scriptsize\color{white}\textbf{$t = 3$}}
        \end{overpic}\hfill
        \includegraphics[width=0.195\linewidth]{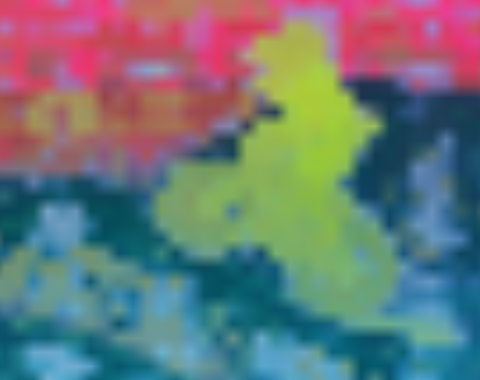}\hfill
        \includegraphics[width=0.195\linewidth]{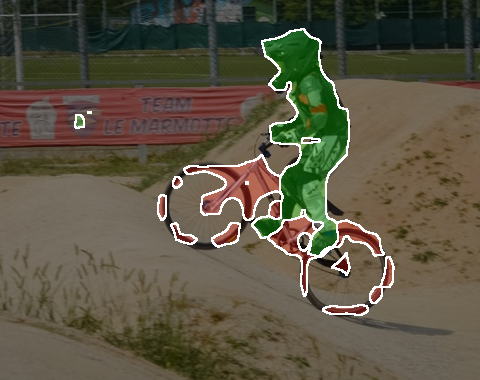}\hfill
        \includegraphics[width=0.195\linewidth]{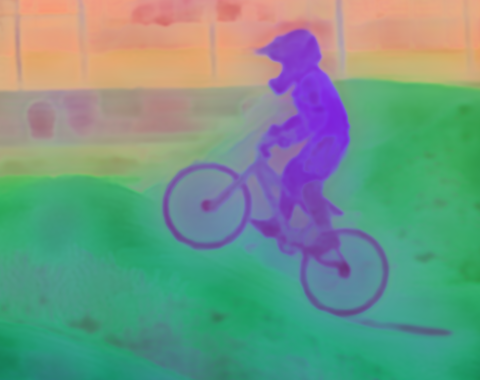}\hfill
        \includegraphics[width=0.195\linewidth]{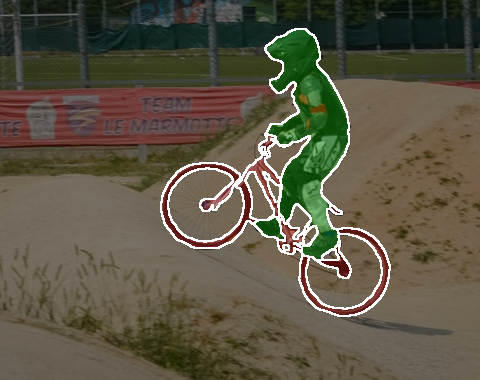}
    \end{subfigure}%

    \vspace{0.2em}

    \begin{subfigure}[t]{\textwidth}
        \centering

        \begin{overpic}[width=0.195\linewidth]{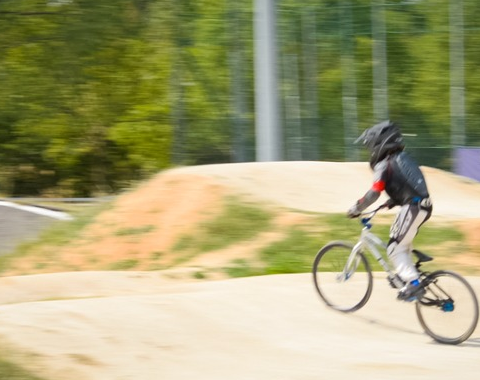}
            \put(72,70){\scriptsize\color{white}\textbf{$t = 17$}}
        \end{overpic}\hfill
        \includegraphics[width=0.195\linewidth]{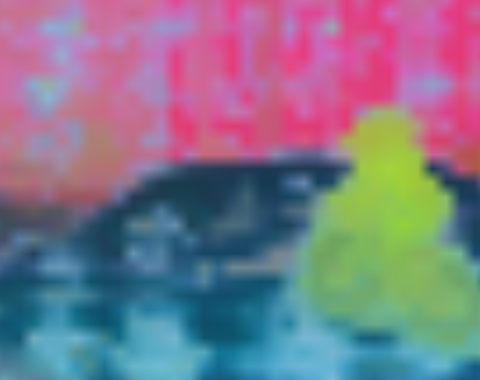}\hfill
        \includegraphics[width=0.195\linewidth]{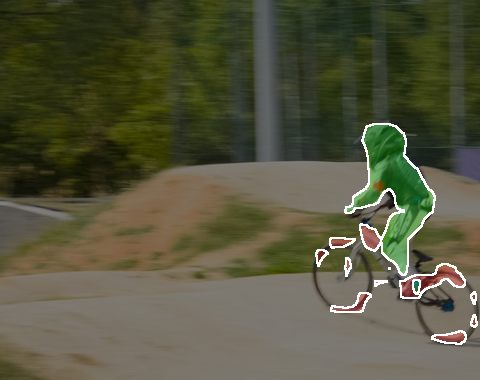}\hfill
        \includegraphics[width=0.195\linewidth]{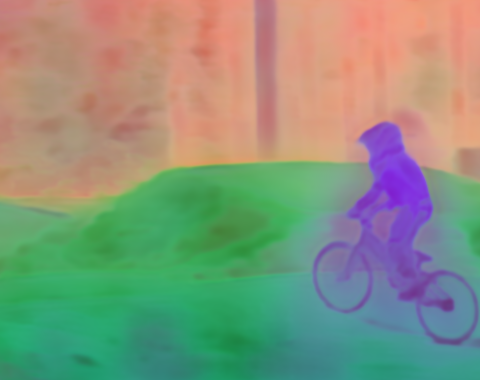}\hfill
        \includegraphics[width=0.195\linewidth]{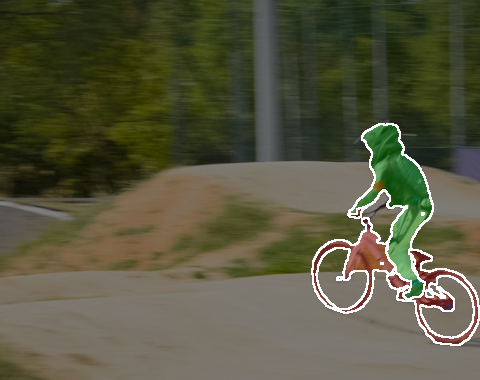}
    \end{subfigure}%

    \caption{\textbf{Qualitative results on VOS.} \oursName provides a high degree of spatial detail and strong temporal consistency, allowing for a high-quality segmentation of fine and dynamic structures (\eg a bicycle).}
    \label{fig:vos}
    \vspace{-0.5em}
\end{figure*}

\subsection{Video object segmentation}
\label{sec:vos}

We evaluate \oursName on the downstream task of video object segmentation.
The goal is to estimate correct segmentation of a video sequence based on the ground-truth annotation provided for the first frame.
The validation sequences feature camera motion as well as motion of one or multiple objects.
As the standard metrics on this benchmark, we report the mean Jaccard index (IoU), $J_m$ and the boundary-based F-score, $F_m$.
Following previous work \cite{Araslanov:2025:FFT}, we test \oursName with two probing methodologies: linear probing and local \textit{k}-NN.

\inparagraph{Linear probing.}
Using the annotation in the first frame, we fit a $1\times 1$ convolution that maps a feature representation to the segmentation logits.
Then, we apply the linear projection to the features extracted from the other frames.

\inparagraph{Local \textit{k}-NN.}
Following previous work \citep{Jabri:2020:CRW,Caron:2021:EPS}, we evaluate \oursName on VOS with autoregressive label propagation.
Given a feature map at timestep $t$, we estimate $k$ nearest neighbours from previous frames in the local window of each pixel.
Since this approach is highly sensitive to the hyperparameters (\eg the size of the local window \etc), we stick to the same hyperparameter setting as in DINO~\cite{Caron:2021:EPS}.

\begin{table*}[t]

\captionof{table}{\textbf{Probing surface normals and semantic segmentation.} We report the probing accuracy on NYUv2 (val) for surface normals in terms of Root Mean Squared Error (RMSE) and inlier ratios for thresholds $\delta_1 = 11.25^\circ$, $\delta_2 = 22.5^\circ$, $\delta_3 = 30^\circ$. We evaluate semantic segmentation on COCO-Stuff (val) and report the mean IoU and the average pixel accuracy (pAcc). \oursName provides a confident boost to the accuracy of all baselines we tested and across all metrics.}
\label{tab:depth_results}

\newcolumntype{B}{S[table-format=2.2]}
\newcolumntype{A}{S[table-format=2.2]}
\newcolumntype{Y}{S[table-format=2.1]}
\newcolumntype{C}[1]{>{\centering\arraybackslash}p{#1}}

\medskip
\footnotesize
\centering
\begin{tabularx}{\linewidth}{Xcp{3em}BAAAp{3em}C{8em}C{8em}}
\toprule
\multirow{2}{*}{Method} & \multirow{2}{*}{Train Data} & & \multicolumn{4}{c}{Surface Normal Estimation (NYUv2)} & & \multicolumn{2}{c}{Semantic Segmentation (COCO-Stuff)} \\
\cmidrule(lr){4-7} \cmidrule(lr){9-10}
 & & & {RMSE $\downarrow$} & {$\delta_1 \uparrow$} & {$\delta_2 \uparrow$} & {$\delta_3 \uparrow$} & & {mIoU $\uparrow$} & {pAcc $\uparrow$} \\
\midrule
DINO2-S14 \citep{Oquab:2023:DINOv2} & LVD$^\ast$ & & 29.71 & 26.91 & 57.72 & 71.29 & & 56.6 & 77.2 \\
\leftPad FlowFeat \citep{Araslanov:2025:FFT} & \quad \leftPad  YT-VOS & & 29.04 & 27.89 & 58.86 & 72.35 & & 58.0 & 78.7 \\
\rowcolor{azure}
\leftPad\oursName (ours)                & \quad \leftPad  YT-VOS & &  \bfseries 28.53 &  \bfseries 31.14 &  \bfseries 61.36 &  \bfseries 74.01  & & \bfseries 59.6 & \bfseries 79.8  \\ 
\midrule
DINO2-B14 \citep{Oquab:2023:DINOv2} & LVD$^\ast$ & & 26.56 & 34.33 & 66.01 & 77.71 & & 58.5 & 78.2 \\
\leftPad FlowFeat \citep{Araslanov:2025:FFT} & \quad \leftPad  Kinetics & & 26.28 & 35.60 & 67.01 & 78.46 & & 60.4 & 79.8 \\
\rowcolor{azure}
\leftPad\oursName (ours)                & \quad \leftPad  YT-VOS & & \bfseries 25.71 & \bfseries 37.13 & \bfseries 68.01 & \bfseries 79.24   & & \bfseries 62.4 & \bfseries 81.2  \\
\midrule
DINO2-L14 \citep{Oquab:2023:DINOv2} & LVD$^\ast$ & & 24.70 & 39.89 & 70.73 & 81.17 & & 58.7 & 78.1 \\
\rowcolor{azure}
\leftPad\oursName (ours)                & \quad \leftPad  YT-VOS & & 24.12 & 40.53 & 71.78 & 82.20 & & 62.8 &  81.1  \\
\rowcolor{azure}
\leftPad\oursName (ours)                 & \:\: \leftPad  Kinetics & & \bfseries 24.04 & \bfseries 40.89 & \bfseries 71.82 & \bfseries 82.34   & & \bfseries 63.3 & \bfseries 81.4  \\
\bottomrule
\end{tabularx}
\vspace{-0.5em}
\end{table*}

\begin{figure*}[t] 
    \centering
    
    \begin{subfigure}[t]{\textwidth}
        \centering
        \begin{minipage}{0.16\linewidth}\centering\footnotesize Input\end{minipage}\hfill
        \begin{minipage}{0.16\linewidth}\centering\footnotesize Ground Truth\end{minipage}\hfill
        \begin{minipage}{0.16\linewidth}\centering\footnotesize DINOv2 (PCA)\end{minipage}\hfill
        \begin{minipage}{0.16\linewidth}\centering\footnotesize DINOv2 (Prediction)\end{minipage}\hfill
        \begin{minipage}{0.16\linewidth}\centering\footnotesize \oursName\ (PCA)\end{minipage}\hfill
        \begin{minipage}{0.16\linewidth}\centering\footnotesize \oursName\ (Prediction)\end{minipage}
        
        \includegraphics[width=0.16\linewidth]{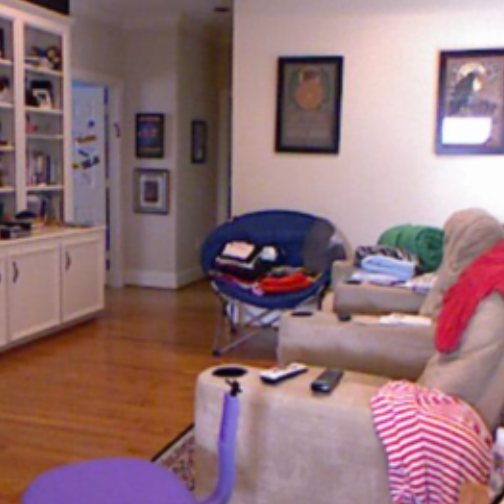}\hfill
        \includegraphics[width=0.16\linewidth]{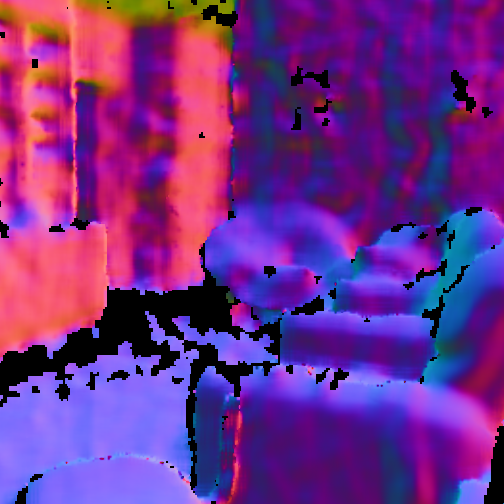}\hfill
        \includegraphics[width=0.16\linewidth]{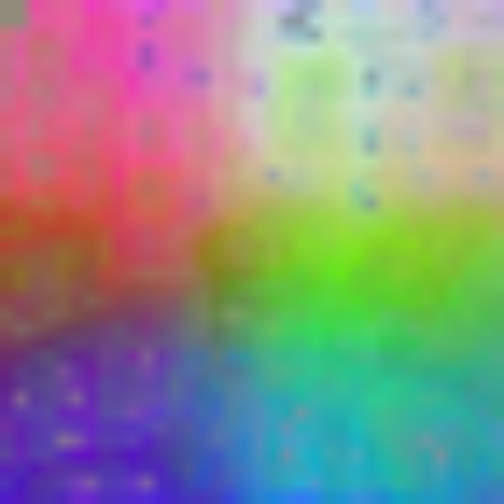}\hfill
        \includegraphics[width=0.16\linewidth]{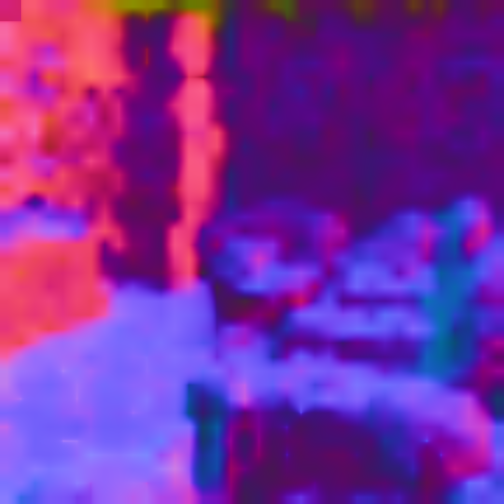}\hfill
        \includegraphics[width=0.16\linewidth]{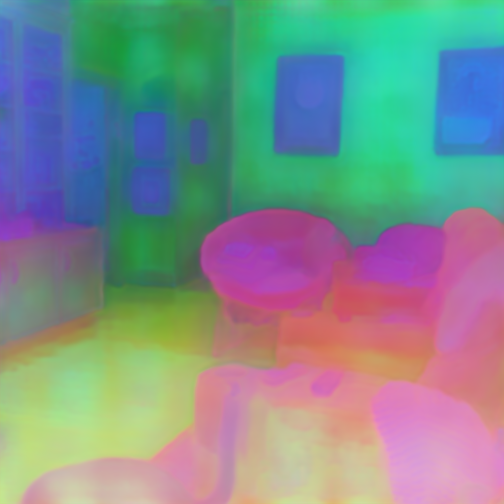}\hfill
        \includegraphics[width=0.16\linewidth]{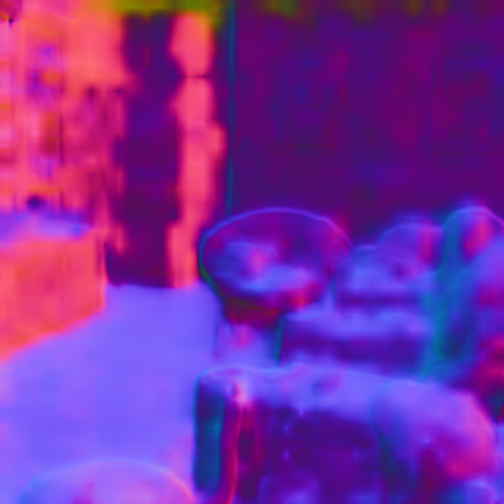}
        \caption{Surface normal estimation (NYUv2)}
        \label{fig:normals}
    \end{subfigure}%

    \begin{subfigure}[t]{\textwidth}
        \centering
        
        \includegraphics[width=0.16\linewidth]{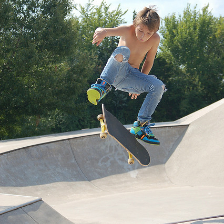}\hfill
        \includegraphics[width=0.16\linewidth]{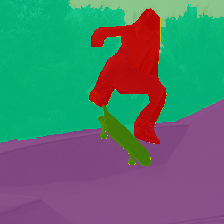}\hfill
        \includegraphics[width=0.16\linewidth]{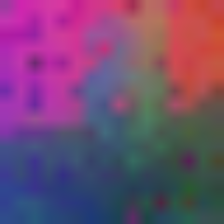}\hfill
        \includegraphics[width=0.16\linewidth]{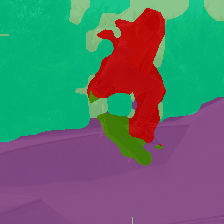}\hfill
        \includegraphics[width=0.16\linewidth]{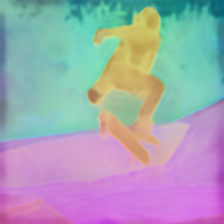}\hfill
        \includegraphics[width=0.16\linewidth]{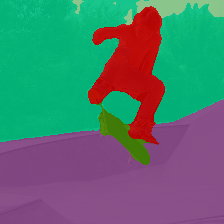}
        \caption{Semantic segmentation (COCO-Stuff)}
        \label{fig:seg}
    \end{subfigure}

    \caption{\textbf{Qualitative results: surface normal estimation and semantic segmentation.} \subref{fig:normals} Despite pre-training on dynamic scenes, \oursName reveals a finer surface structure than the baseline (DINOv2 \cite{Oquab:2023:DINOv2} / ViT-B14) in predominantly static scenes. \subref{fig:seg} \oursName enables more accurate semantic segmentation, providing more accurate semantic boundaries and background detail.}
    \label{fig:segnormals}
    \vspace{-0.5em}
\end{figure*}

\inparagraph{Results.}
\cref{exp:vos} summarises the results of the linear probing (left) and the local \textit{k}-NN mask propagation (right).
We observe a substantial improvement of the VOS accuracy across all baselines and state-of-the-art methods.
For example, \oursName based on DINO2-S14 outperforms the baseline by almost $10~\%$ $\mathcal{JF}$ (linear probing), and surpasses LoftUp by $4.3~\%$ $\mathcal{JF}$.
This is remarkable, since \oursName did not use mask supervision\emdash in contrast to LoftUp, which leveraged SAM~\cite{Kirillov:2023:SAM} for training.
The improvement also translates to larger models.
Based on DINO2-B14, \oursName outperforms the recent FlowFeat~\cite{Araslanov:2025:FFT}, by $4.7~\%$ and $5.2~\%$ $\mathcal{JF}$ with linear probing and \textit{k-}NN, respectively.
Additionally, we trained the largest \oursName variant based on DINO2-L14 on Kinetics, which is significantly larger than YouTube-VOS, though is substantially less well curated.
\oursName scales well to this setting and improves the previous results from YouTube-VOS pre-training for both probing methodologies.
As shown in \cref{fig:vos}, \oursName excels at segmenting fine-grained structures (\eg a bicycle) owing to the spatial detail in the feature maps.

In summary, \oursName yields semantically aware, temporally consistent feature maps, confidently surpassing previous work across all backbones and probing techniques.

\subsection{Surface normal estimation}
\label{sec:snorm}

Following previous work \citep{Banani:2024:Probe3D}, we use NYUv2 dataset \citep{Silberman:2012:ECCV12} for estimating normals per pixel.
We train a linear probe to map the feature representation to a 3D normal vector.
For the high-resolution feature grids of \oursName (and the recent work FlowFeat~\cite{Araslanov:2020:SSS}), we implement the linear mapping using a convolutional layer with a $5 \times 5$ kernel.
We augment the 3D representation from our decoder with the low-resolution output from the encoder, obtained by a $1 \times 1$ convolution and bilinearly upsampled to the original image resolution.

\cref{tab:depth_results} (left section) reports the results in terms of four standard metrics: root mean squared error (RMSE) and the fraction of pixels with the angular error below three pre-defined thresholds $\delta_1 = 11.25^\circ$, $\delta_2 = 22.5^\circ$, $\delta_3 = 30^\circ$.
\oursName outperforms all baselines and the recent FlowFeat across all metrics with a confident margin.
We observe that the surface normal accuracy significantly improves with model capacity.
For example, RMSE of DINO2-L14 is lower than that of DINO2-S14 by around $5$ points.
Despite the stronger baseline, \oursName improves the normal accuracy further.
By training \oursName with DINO2-L14 on the larger Kinetics, we achieve a modest, though still tangible boost.

Observing the visual results in \cref{fig:normals}, we note a remarkable degree of detail in the feature maps yielded by \oursName.
Furniture and household items are especially prominent in the representation.
This is surprising, since \oursName was trained on dynamic scenes with a strong bias for outdoor scenes and dynamic objects, not static environments.

Overall, the results suggest that \oursName encodes geometric scene properties, while also embedding the semantic structure of complex scenes,
which we analyse further next.

\subsection{Semantic segmentation}
\label{sec:sseg}

We use COCO-Stuff \citep{Caesar:2018:COCO} for probing experiments on semantic segmentation.
The task is to predict categorical distribution for each image pixel.
Following previous work \citep{Fu:2024:FeatUp}, we use the coarse set annotation with $C=27$ semantic categories.
We use the train split of COCO-Stuff for training the probe.
Adapting the strategy from FlowFeat~\cite{Araslanov:2025:FFT}, \oursName uses an attention-based probe.
Specifically, we pass $C$ query tokens via cross-attention through the spatial feature grid to obtain class prototypes.
For efficiency, we downsample the high-resolution feature grid in the cross-attention block.
The logits for each pixel derive from a scalar product of the class prototypes and the original feature map. 
We complement the high-resolution logits obtained from the decoder by a sum with the low-resolution logits from the encoder, which we obtain via a concurrently trained $1\times 1$ convolution.

In \cref{tab:depth_results} (right section), we report the results on the validation set of COCO-Stuff in terms of the mean IoU (\textit{mIoU}) and the average pixel accuracy (\textit{pAcc}).
\oursName consistently and notably improves the segmentation quality across all baselines and the recent FlowFeat~\cite{Araslanov:2025:FFT}.
The absolute improvement tends to increase with the model size.
For example, \oursName boosts the \textit{mIoU} of DINO2-S14 by $+3.0~\%$, whereas for DINO2-L14 the benefit amounts to $+4.1~\%$.
This observation suggests that \oursName succeeds at leveraging more complex patterns available in the encoder network (via skip connections).
To test the scalability further, we evaluate a \oursName variant trained on Kinetics.
Here, we observe a tangible improvement across both \textit{mIoU} and \textit{pAcc} in comparison to the small-scale pre-training ($+0.5~\%$ / $+0.3~\%$, respectively).
This suggests that \oursName has the capacity to scale its semantic representation from uncurated videos.

From the qualitative results in \cref{fig:seg}, we note a dramatic improvement of the feature quality in terms of the spatial detail and semantic granularity.
The improvement is particularly pronounced at the semantic boundaries, for both static (\eg picture frame) and dynamic elements (\eg skateboard). 

\begin{table}[t]
\captionof{table}{\textbf{Probing segmentation: ADE20K and zero-shot COCO-Stuff.} \subref{tab:addseg-a} Using linear probing, we evaluate \oursName on ADE20K. \subref{tab:addseg-b} In the zero-shot setting, we train a linear probe initialised from the text embeddings, extracted from CLIP~\cite{Radford:2021:LTV}, of seen classes in COCO-Stuff and test on 15 unseen categories, reporting mIoU.}
\label{tab:addseg}

\newcolumntype{A}{S[table-format=2.1]}
\newcolumntype{C}[1]{>{\centering\arraybackslash}p{#1}}

\medskip
\footnotesize
\centering
\begin{tabularx}{\linewidth}{XAAAA}
\toprule
\multirow{2}{*}{Method} & {\tabmark{tab:addseg-a}{ADE20K}} & \multicolumn{3}{c}{\tabmark{tab:addseg-b}{Zero-shot: COCO-Stuff}} \\
\cmidrule(lr){2-2} \cmidrule(lr){3-5}
& {mIoU} & {Seen} & {Unseen} & {Harmonic} \\
\midrule
DINO2-S14 \citep{Oquab:2023:DINOv2} & 43.5 & 30.1 & 17.0 & 21.8 \\
\rowcolor{azure}
\leftPad\oursName (ours) & \bfseries 45.1 & \bfseries 32.8 &  \bfseries 20.1 &  \bfseries 24.9 \\ 
\midrule
DINO2-B14 \citep{Oquab:2023:DINOv2} & 45.5 & 32.6 & 19.0 & 24.1 \\
\rowcolor{azure}
\leftPad\oursName (ours) & \bfseries 47.5 & \bfseries 34.6 &  \bfseries 22.7 &  \bfseries 27.4 \\ 
\bottomrule
\end{tabularx}
\end{table}

\inparagraph{Semantic segmentation on ADE20K \cite{Zhou:2017:ADE}.}
COCO-Stuff is substantially larger and more diverse than ADE20k (100K \text{vs} 20K training images) and enables a large-scale evaluation focused on robustness across diverse scenes.
Nevertheless, we additionally verify the benefit of LILA on ADE20k using DINO2 ViT-S14 and ViT-B14 as baselines in \cref{tab:addseg}\subref{tab:addseg-a}.
We observe a consistent improvement (\eg \num{43.5}\% $\to$ \num{45.1}\% mIoU using DINO-S14) on the ADE20K benchmark as well.

\inparagraph{Zero-shot segmentation.}
We design a probing experiment to evaluate zero-shot segmentation on COCO-Stuff.
Specifically, we initialise the probe using CLIP embeddings \cite{Radford:2021:LTV}, split into seen and unseen categories.
While training the probe on the seen categories, we report the results on a held-out set containing both seen and unseen classes.
The results in \cref{tab:addseg}\subref{tab:addseg-b} show that \oursName also captures visual patterns of unseen classes, outperforming the baseline on all metrics.

\subsection{Further analysis}

We verify the design choices to supervise and train \oursName
and report the results on the benchmarks from \cref{sec:vos,sec:snorm,sec:sseg}.

\inparagraph{Cue modalities, \cf \cref{tab:ablation_cues}.}
Recall that we use three cues in training: self-distillation (SD), depth and optical flow.
We investigate the degree to which these cues affect the downstream accuracy.
First, we remove the geometric and motion cues, which leads to a significant drop in VOS-KNN accuracy ($-5.3~\%$ $\mathcal{JF}$).
Self-distillation, depth and optical flow improve \oursName by the tangible $1.3~\%$ $\mathcal{JF}$, $1.3~\%$ $\mathcal{JF}$ and $1.4~\%$ $\mathcal{JF}$, respectively.
Overall, these results suggest a strong synergistic effect: the downstream accuracy improves markedly with diversity of the cues provided for supervision. 

\begin{table}[t]

    \caption{\textbf{Ablation study: cue modalities.} We train \oursName with varying modalities in the cue maps and report the accuracy on VOS (DAVIS 2017), surface normal estimation (NYUv2) and semantic segmentation (COCO-Stuff).}
    \label{tab:ablation_cues}

\newcolumntype{Y}{>{\centering\arraybackslash}X}
\newcolumntype{D}{S[table-format=2.1]}
\newcolumntype{N}{S[table-format=2.2]}
\newcolumntype{C}{S[table-format=2.1]}

\footnotesize
\centering
\setlength{\tabcolsep}{3pt}

\begin{tabularx}{\linewidth}{YYY D N C}
\toprule
\multicolumn{3}{c}{Cue Modalities}

& \multicolumn{1}{c}{\textbf{DAVIS 2017}} 
& \multicolumn{1}{c}{\textbf{NYUv2}} 
& \multicolumn{1}{c}{\textbf{COCO-Stuff}} \\

\cmidrule(lr){1-3} \cmidrule(lr){4-4} \cmidrule(lr){5-5} \cmidrule(lr){6-6}

SD\textsuperscript{*} &  Depth &  Flow 
& \multicolumn{1}{c}{\scriptsize LP / KNN $\mathcal{JF} \uparrow$} 
& \multicolumn{1}{c}{RMSE\(\downarrow\)} 
& \multicolumn{1}{c}{mIoU\(\uparrow\)} \\

\midrule
$\checkmark$ & &  & {66.9 / 68.6} & {28.61} & {59.3} \\
& $\checkmark$ & $\checkmark$ & {67.2 / 72.6} & {29.06} & {58.7} \\
$\checkmark$ &  & $\checkmark$ & {67.0 / 72.6} & 28.64 & 59.5 \\
$\checkmark$ & $\checkmark$ &  & {{\bfseries 69.1} / 72.5} & {\bfseries 28.49} & {59.3} \\
\addlinespace
\rowcolor{azure}
$\checkmark$ & $\checkmark$ & $\checkmark$  & {68.6 / {\bfseries 73.9}} &  28.53 & \bfseries 59.6 \\
\bottomrule
\multicolumn{6}{@{}l}{\footnotesize \textsuperscript{*}\,SD = Self-Distillation with PAMR} \\
\end{tabularx}
\vspace{-0.5em}
\end{table}

\begin{table}[t]

\caption{\textbf{Ablation study: training components}. We switch off \oursName's training components to assess their downstream impact on VOS (DAVIS 2017), surface normal estimation (NYUv2) and semantic segmentation (COCO-Stuff).}
\label{tab:ablation_components}

\newcolumntype{D}{S[table-format=2.1]}
\newcolumntype{N}{S[table-format=2.2]}
\newcolumntype{C}{S[table-format=2.1]}

\footnotesize
\centering
\setlength{\tabcolsep}{3pt}

\begin{tabularx}{\linewidth}{c X D N C}
\toprule

&
Baselines
& \multicolumn{1}{c}{\textbf{DAVIS 2017}} 
& \multicolumn{1}{c}{\textbf{NYUv2}} 
& \multicolumn{1}{c}{\textbf{COCO-Stuff}} \\

\cmidrule(lr){3-3} \cmidrule(lr){4-4} \cmidrule(lr){5-5}

&
w/ DINO2-S14
& \multicolumn{1}{c}{\scriptsize LP / KNN $\mathcal{JF} \uparrow$} 
& \multicolumn{1}{c}{RMSE\(\downarrow\)} 
& \multicolumn{1}{c}{mIoU\(\uparrow\)} \\

\midrule
\textit{(A)} & ERM distillation & {63.2 / 61.1} & 28.86 & 58.4 \\
\textit{(B)} & \xmark~PAMR & {67.3 / 71.9} & 28.58 & 59.1 \\
\textit{(C)} & \xmark~cropping & {66.0 / 72.4} & 28.93 & 59.3 \\
\textit{(D)} & \xmark~temporal sampling & {{\bfseries 69.3} / 72.4} & 28.74 & 59.3 \\
\textit{(E)} & \xmark~edge loss & {68.1 / 72.9} & 28.67 & 59.4 \\
\addlinespace
\rowcolor{azure}
& \oursName (Full) & {68.6 / {\bfseries 73.9}} & \bfseries 28.53 & \bfseries 59.6 \\
\bottomrule

\end{tabularx}
\vspace{-0.5em}
\end{table}

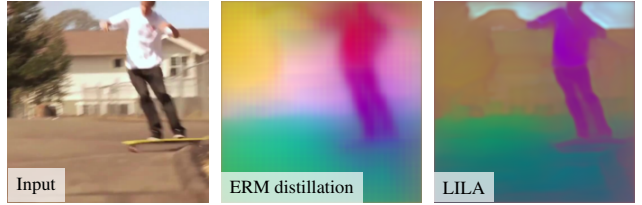
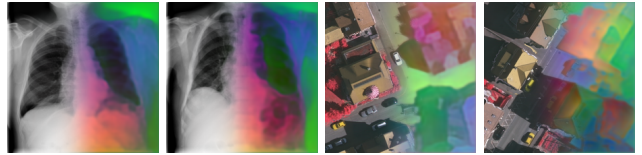
\begin{figure}[t]
    \centering

    \begin{subfigure}{\linewidth}
        \centering
        \begin{minipage}[t]{0.32\linewidth}
            \centering
            \begin{tikzpicture}
                \node[inner sep=0] (img) {
                    \includegraphics[width=\linewidth]{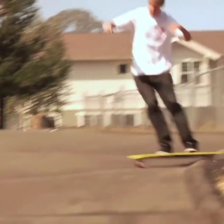}
                };
                \node[
                    anchor=south west,
                    font=\scriptsize,
                    fill=white,
                    fill opacity=0.7,
                    text opacity=1
                ] at (img.south west) {Input};
            \end{tikzpicture}
        \end{minipage}
        \hfill
        \begin{minipage}[t]{0.32\linewidth}
            \centering
            \begin{tikzpicture}
                \node[inner sep=0] (img) {
                    \includegraphics[width=\linewidth]{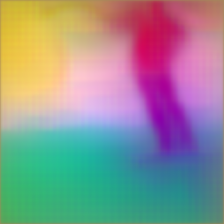}
                };
                \node[
                    anchor=south west,
                    font=\scriptsize,
                    fill=white,
                    fill opacity=0.7,
                    text opacity=1
                ] at (img.south west) {ERM distillation};
            \end{tikzpicture}
        \end{minipage}
        \hfill
        \begin{minipage}[t]{0.32\linewidth}
            \centering
            \begin{tikzpicture}
                \node[inner sep=0] (img) {
                    \includegraphics[width=\linewidth]{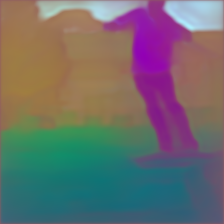}
                };
                \node[
                    anchor=south west,
                    font=\scriptsize,
                    fill=white,
                    fill opacity=0.7,
                    text opacity=1
                ] at (img.south west) {LILA};
            \end{tikzpicture}
        \end{minipage}

        \caption{Comparison to ERM distillation}
        \label{fig:erm_distill}
    \end{subfigure}

    \vspace{0.5em}

    \begin{subfigure}{\linewidth}
        \begin{minipage}{\columnwidth}
            \centering
            \includegraphics[width=0.24\columnwidth]{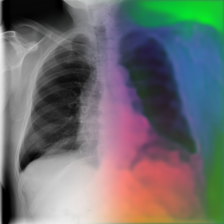}
            \hfill
            \includegraphics[width=0.24\columnwidth]{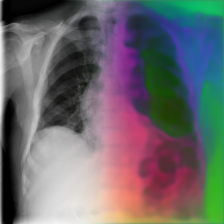}
            \hfill
            \includegraphics[width=0.24\columnwidth]{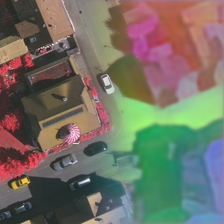}
            \hfill
            \includegraphics[width=0.24\columnwidth]{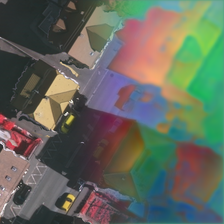}
        \end{minipage}
        
        \caption{Out-of-domain results}
        \label{fig:ood_examples}
    \end{subfigure}

    \caption{\textbf{Further qualitative analysis}. \subref{fig:erm_distill} A qualitative comparison between \oursName and ERM distillation\emdash a distillation baseline based on the external cues without linear in-context learning. \oursName yields noticeably sharper feature maps, suggesting that linear in-context learning is effective at handling inherent noise in the external cues. \subref{fig:ood_examples} We test \oursName on out-of-domain images\emdash here, chest X-rays and aerial imagery. Although \oursName produces plausible representations for chest X-rays, its representations for aerial images are confounded by shadows, revealing a limitation.}
    \label{fig:addon_qualitative}
\end{figure}

\inparagraph{Training components, \cf \cref{tab:ablation_components}.}
We assess the role of individual components in the \oursName training framework.
First, $\textit{(A)}$ we trained the DPT head to directly predict the same external cues (depth, flow and self-distillation) from a single frame. 
This ``ERM distillation'' baseline yields consistently lower downstream accuracy than LILA.
Including or excluding optical flow had no significant effect.
Since the cues are identical, this performance gap is directly attributable to our in-context training strategy, instead of the external cues.
Recall that the external cues \emph{do not constitute a clean training signal}, but rather noisy observations.
\oursName's cross-frame formulation suppresses frame-specific, temporally unstable noise components.
By contrast, standard distillation (ERM) absorbs the noise in expectation.
\cref{fig:erm_distill} illustrates the resulting difference in representations.
Second, $\textit{(B)}$ we remove the refinement of the encoder representation in self-distillation with PAMR.
A significant drop in the VOS-KNN accuracy ($-2.0~\%$ $\mathcal{JF}$) indicates an important role of PAMR in the training process.
$\textit{(C)}$ We remove cropping to synthesise queries in our in-context framework.
Here, the accuracy deteriorates (by $1.5~\%\; \mathcal{JF}$).
Therefore, cropping makes learning effective by creating diversity of queries.
$\textit{(D)}$ By removing temporal sampling, we constrain \oursName training to operate on a single image.
Though the drop in VOS-KNN accuracy is notable ($-1.5~\%\; \mathcal{JF}$), the overall downstream accuracy suggests that \oursName can benefit further from image-based pre-training.
Finally, $\textit{(E)}$ we remove the edge loss (\cf \cref{eq:grad-loss}).
The drop in VOS-KNN accuracy of $-1.0~\%\; \mathcal{JF}$, suggests that penalising at boundaries promotes semantic structures.
These VOS-KNN trends also hold for surface normal estimation and semantic segmentation.

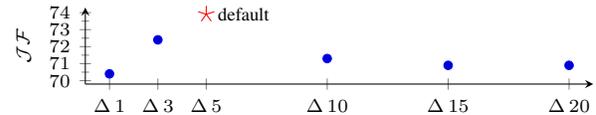
\begin{figure}
\begin{tikzpicture}
\begin{axis}[
  width=\columnwidth,
  height=2.6cm,
  enlargelimits=false,
  ylabel={\scriptsize \(\mathcal{JF}\)},
  xmin=0, xmax=21,
  ymin=69.8, ymax=74.3,
  xtick={1,3,5,10,15,20},
  ytick distance=1,
  tick label style={font=\scriptsize},
  xticklabel={$\mathord{\Delta}\,\pgfmathprintnumber{\tick}$},
  axis lines=left,
  major grid style={line width=.2pt},
  minor grid style={line width=.1pt},
  minor tick num=1,
  xlabel style={draw=none},
  ylabel style={draw=none},
]

\addplot+[
  only marks,
  mark=*,
  mark size=1.6pt
] coordinates {
  (1,70.4)
  (3,72.4)
  (10,71.3)
  (15,70.9)
  (20,70.9)
};

\node[anchor=west,font=\scriptsize]
  at (axis cs:5.1,73.9) {default};

\addplot+[
  only marks,
  mark=star,
  mark size=3.2pt
] coordinates {
  (5,73.9)
};

\end{axis}
\end{tikzpicture}
\caption{\textbf{The temporal gap trade-off.} We report $\mathcal{JF}$ on VOS (DAVIS 2017) with \text{k}-NN inference for different \oursName models trained with varying sizes of the sampling window $\Delta$. }
\label{fig:temp_ablation}
\vspace{-0.4em}
\end{figure}

\inparagraph{The temporal gap $\Delta$, \cf \cref{fig:temp_ablation}.}
There is a clear trade-off, illustrated by $\mathcal{JF}$ with DINO2-S14 using local \textit{k}-NN.
Small $\Delta$ simplifies the task, resulting in weaker representations.
As $\Delta$ increases, cue predictability becomes more challenging. Accuracy degrades gracefully for large $\Delta$.

\inparagraph{Limitations.}
\oursName relies on cue maps, which derive from state-of-the-art off-the-shelf models (SEA-RAFT~\cite{Wang:2024:SEA} for optical flow; DepthAnythingV2~\cite{Yang:2024:DAv2} for depth).
Therefore, the scope of applicability is currently limited to the realm, where the predicted depth and optical flow are plausible.
\cref{fig:ood_examples} visualises some out-of-domain examples from the medical and aerial domains.
We find that \oursName yields reasonable representations for chest X-rays, whereas representations of aerial scenes are not always meaningful due to shadows.

\section{Conclusion}

We presented linear in-context learning, an effective training strategy for encoder-decoder networks.
Learned from videos and imperfect cue maps, \oursName effectively embeds temporal, semantic and geometric aspects into its pixel-level representation.
We evaluated  \oursName across three benchmarks and demonstrated a substantial and consistent boost of the encoder representation across all baselines and metrics.

{
    \small

    \makeatletter
    \arxivciteanchor{main-He:2022:MAA}
    \arxivciteanchor{main-Simeoni:2025:DINOv3}
    \arxivciteanchor{main-Araslanov:2020:SSS}
    \arxivciteanchor{main-Oquab:2023:DINOv2}
    \arxivciteanchor{main-Radford:2021:LTV}
    \arxivstopcitationwrites
    \makeatother
}

\clearpage
\pagenumbering{roman}

\arxivsettitle{Featurising Pixels from Dynamic 3D Scenes with Linear In-Context Learners\\[2mm]
\large -- Supplemental Material --}
\arxivsetauthor{\arxivauthors}
\arxivmaketitle

\appendix
\setcounter{table}{5}
\setcounter{equation}{6}

\section{Backbone generalisation}

We train \oursName with different backbone families.
Concretely, we experiment with a Masked Autoencoder (MAE-B16) \cite{main-He:2022:MAA}, a DINOv2 variant equipped with registers (DINO2-B14-Reg) \cite{Darcet:2024:VTN} and DINOv3 \cite{main-Simeoni:2025:DINOv3}.
Here, we focus on the ViT-B architecture.
\cref{tab:backbones} reports the results on VOS with linear probing.
As expected, \oursName provides improved VOS accuracy across the backbone models.
For example it improves the MAE-B16 model by $9.4~\%$ $\mathcal{JF}$.

\section{Practical details}

Training \oursName is computationally inexpensive, a benefit we achieve through several practical considerations.
For example, we downsample the context cue grid, $G_\text{context}$, by a factor of $7$.
This measure significantly improves the efficiency of ridge regression without effecting the downstream performance.
We train all models on YouTube-VOS with a batch size of 32 on a single A100 GPU with 40GB of memory.
The training typically reaches convergence after 30K iterations, which translates to approximately 24 hours.

\inparagraph{PAMR.}
We apply PAMR~\cite{main-Araslanov:2020:SSS} in a coarse-to-fine fashion.
Specifically, we define an image pyramid with three levels corresponding to the downsampling ratios of $4$, $2$ and $1$ (\ie the original resolution).
We define the PAMR parameters identically at each level.
The local affinity kernel consists of three $3 \times 3$ kernels with dilation ratios of $1$, $3$ and $5$.
At the coarsest resolution, the refinement runs for $20$ iterations, and we reduce this number by a factor of $2$ for each consecutive, finer resolution.
This results in a total of $35$ refinement iterations.
Notably, this coarse-to-fine strategy is approximately four times more efficient than running all 35 iterations at the original image resolution.

\begin{table}[t]

\captionof{table}{\textbf{Generalisation across other backbones.} We report the probing accuracy by pre-training \oursName with diverse backbones: Masked Autoencoder, DINOv2 with registers and DINOv3.}
\label{tab:backbones}

\newcolumntype{B}{S[table-format=2.2]}
\newcolumntype{A}{S[table-format=2.2]}
\newcolumntype{Z}{S[table-format=2.2]} 
\newcolumntype{Y}{S[table-format=2.1,table-number-alignment=center]} 
\newcolumntype{Q}{S[table-format=2.1,input-symbols = ()]} 

\medskip
\footnotesize
\centering
\begin{tabularx}{\linewidth}{Xcp{1em}YYY}
\toprule
\textbf{Method} & Train Data & & $\mathcal{JF}$ & $\mathcal{J}_m$  & $\mathcal{F}_m$ \\
\midrule
MAE-B16 \citep{main-Oquab:2023:DINOv2} & LVD$^\ast$ & & 44.2 & 41.7 & 46.7 \\
\rowcolor{azure}
\leftPad\oursName (ours)                & \quad \leftPad  YT-VOS & & \bfseries 53.6 & \bfseries 50.4 & \bfseries 56.8 \\ 
\midrule
DINO2-B14-Reg \citep{main-Oquab:2023:DINOv2} & LVD$^\ast$ & & 61.6 & 59.1 & 64.2 \\
\rowcolor{azure}
\leftPad\oursName (ours)                & \quad \leftPad  YT-VOS & & \bfseries 68.4 & \bfseries 64.7 & \bfseries 72.1 \\ 
\midrule
DINO3-B16 \citep{main-Oquab:2023:DINOv2} & LVD$^\ast$ & & 63.3 & 60.9 & 65.8 \\
\rowcolor{azure}
\leftPad\oursName (ours)                & \quad \leftPad  YT-VOS & & \bfseries 64.8 & \bfseries 62.0 & \bfseries 67.6 \\ 
\bottomrule
\end{tabularx}
\end{table}

\section{Zero-shot linear probing}
We evaluate \oursName under the standard zero-shot semantic segmentation protocol on COCO-Stuff, following the seen/unseen split introduced in prior work \cite{Xian:2019:SPN}.
After partitioning the dataset categories into seen and unseen classes, we train the probe using only pixels annotated with seen labels, while evaluating generalization on the 15 held-out unseen classes. 
During this evaluation, we keep the pre-trained LILA representation frozen and optimise only a lightweight pixel-level probe.
Concretely, given frozen encoder and decoder feature maps $f^{\mathrm{enc}}$ and $f^{\mathrm{dec}}$, we learn shallow projection heads $P_{\mathrm{enc}}$ and $P_{\mathrm{dec}}$ that map both branches into a common semantic embedding space.
We normalise the projected features with channel-wise $\ell_2$-normalisation and classify them with a fixed linear classifier, in which the weights are provided by the class text embeddings (\eg, CLIP text features) \cite{main-Radford:2021:LTV}. 
This yields per-pixel cosine-similarity logits
\begin{equation}
\begin{aligned}
  \ell_c^{\mathrm{enc}}(u) &=\left\langle \frac{P_{\mathrm{enc}}(f^{\mathrm{enc}}(u))}{\|P_{\mathrm{enc}}(f^{\mathrm{enc}}(u))\|_2},\, t_c \right\rangle,\qquad\\
\ell_c^{\mathrm{dec}}(u) &=\left\langle \frac{P_{\mathrm{dec}}(f^{\mathrm{dec}}(u))}{\|P_{\mathrm{dec}}(f^{\mathrm{dec}}(u))\|_2},\, t_c \right\rangle,  
\end{aligned}
\end{equation}
where $t_c$ denotes the text embedding of class $c$. The final prediction combines both branches,
\begin{equation}
\ell_c(u)=\exp(\gamma)\big(\alpha_{\mathrm{enc}}\,\ell_c^{\mathrm{enc}}(u)+\alpha_{\mathrm{dec}}\,\ell_c^{\mathrm{dec}}(u)\big),
\end{equation}
with learnable logit scale $\gamma$. We use both encoder and decoder representations because the decoder feature maps are expected to provide dense cues that are complementary to the encoder representation.

The probe is trained with a pixel-wise cross-entropy loss on labeled pixels,
\begin{equation}
\mathcal{L}_{\mathrm{CE}}
=
\frac{1}{|\Omega_{\mathrm{lab}}|}
\sum_{u\in\Omega_{\mathrm{lab}}}
\mathrm{CE}\!\left(p(u), y(u)\right),
\end{equation}
where $\Omega_{\mathrm{lab}}$ excludes pixels from the unseen categories.
In addition, for these ignored pixels $\Omega_{\mathrm{ign}}$, we apply a negative regulariser that suppresses probability mass assigned to the seen classes,
\begin{equation}
\mathcal{L}_{\mathrm{neg}}
=
\frac{1}{|\Omega_{\mathrm{ign}}|}
\sum_{u\in\Omega_{\mathrm{ign}}}
\left(\sum_{c\in\mathcal{C}_{\mathrm{seen}}} p_c(u)\right)^2 .
\end{equation}
When ignored pixels are present, the training objective is
\begin{equation}
\mathcal{L}_\text{zero-shot}=0.1\,\mathcal{L}_{\mathrm{CE}}+\mathcal{L}_{\mathrm{neg}},
\end{equation}
and otherwise we optimize only $\mathcal{L}_{\mathrm{CE}}$. 
This regulariser discourages the probe from collapsing unlabelled regions into seen categories and leaves room for transfer to unseen classes. 
At test time, we apply the same frozen text-based classifier without further adaptation.
As a result, the segmentation accuracy on the held-out classes reflects the zero-shot transfer ability of the frozen LILA features and, in particular, the complementarity of the learned decoder representation (\ie \oursName) with that of the encoder.

\end{document}